\documentclass{ar-1col-S2O}

\usepackage[comma]{natbib}
\usepackage{url}
\jname{Xxxx. Xxx. Xxx. Xxx.}
\jvol{AA}
\jyear{YYYY}
\doi{10.1146/((please add article doi))}

\usepackage{graphicx}

\usepackage[hidelinks, colorlinks=blue,citecolor=blue]{hyperref} 
\usepackage{url}
\usepackage{booktabs}
\usepackage{xcolor}
\usepackage{amsthm,amsmath,amsfonts,amssymb}
\usepackage{mathtools}
\usepackage{color}
\usepackage{subcaption}
\usepackage[flushleft]{threeparttable}
\usepackage[shortlabels]{enumitem}
\usepackage{multirow}
\usepackage{pifont} 
\newcommand{\cmark}{\ding{51}} 
\newcommand{\xmark}{\ding{55}} 

\newtheorem{theorem}{Theorem}

\newtheorem{assumption}[theorem]{Assumption}

\newtheorem{lemma}{Lemma}[section]

\newtheorem{conjecture*}{Conjecture}
\theoremstyle{remark}

\begin{document}

\markboth{Cheng, Dong, Xie}{Deep spatio-temporal point processes}

\title{Deep Spatio-temporal Point Processes: Advances and New Directions\footnote{Authors are listed alphabetically.}}

\author{Xiuyuan Cheng,$^1$ Zheng Dong,$^2$ and Yao Xie$^3$
\affil{$^1$Department/Institute, University, City, Country, Postal code; email: author@email.edu}
\affil{$^2$Department/Institute, University, City, Country, Postal code}
\affil{$^3$Department/Institute, University, City, Country, Postal code}}

\begin{abstract}
Spatio-temporal point processes (STPPs) model discrete events distributed in time and space, with important applications in areas such as criminology, seismology, epidemiology, and social networks. Traditional models often rely on parametric kernels, limiting their ability to capture heterogeneous, nonstationary dynamics. Recent innovations integrate deep neural architectures—either by modeling the conditional intensity function directly or by learning flexible, data-driven influence kernels—substantially broadening their expressive power. This article reviews the development of the deep influence kernel approach, which enjoys statistical explainability—since the influence kernel remains in the model to capture the spatiotemporal propagation of event influence and its impact on future events—while also possessing strong expressive power, thereby benefiting from both worlds. We explain the main components in developing deep kernel point processes, leveraging tools such as functional basis decomposition and graph neural networks to encode complex spatial or network structures, as well as estimation using both likelihood-based and likelihood-free methods, and address computational scalability for large-scale data. We also discuss the theoretical foundation of kernel identifiability. Simulated and real-data examples highlight applications to crime analysis, earthquake aftershock prediction, and sepsis prediction modeling, and we conclude by discussing promising directions for the field.
\end{abstract}

\begin{keywords}
Spatio-temporal point processes,
Deep learning,
Functional low-rank representation,
Non-stationary kernels,
Influence function estimation,
Event prediction
\end{keywords}

\maketitle


\tableofcontents

\section{Introduction}

Spatio-temporal event data—discrete events that occur at specific times and locations—have become central to numerous research areas, including criminology (e.g., crime incidents in urban settings), seismology (e.g., earthquakes and aftershocks), epidemiology (e.g., disease outbreaks), and social network analysis. In these applications, the data tend to exhibit an excitation pattern -- that one event can trigger subsequent events in nearby locations and times. For example, in criminology, there is the so-called ``broken window theory'', such that a crime in one neighborhood can increase the likelihood of related crimes occurring in surrounding areas. Understanding such patterns is important for scientific inquiry and practical applications, such as forecasting future occurrences and clarifying event-to-event causal relationships.

Spatio-temporal point processes (STPPs) provide a powerful statistical framework for modeling discrete events. They have been successfully applied to phenomena such as earthquake occurrences \citep{ogata1988statistical, ogata2003modelling, zhuang2004analyzing, Kumazawa2014}, the spread of infectious diseases \citep{meyer2012space, Meyer2014, schoenberg2019recursive, dong2023non}, and crime dynamics \citep{mohler2011self, mohler2014marked, reinhart2018self}. Mathematically, STPPs represent each event as a pair \((t, s)\), where \(t\) is the event time and \(s\) its spatial location or mark. A primary object of interest is the conditional intensity function \(\lambda(t, s)\), which indicates how the probability of a new event depends on the history of prior events. Modern datasets often include additional high-dimensional marks, leading to correspondingly high-dimensional conditional intensities.

Classical models typically adopt a self-exciting structure for \(\lambda(t, s)\), following the seminal Hawkes process framework \citep{hawkes1971spectra}, in which a baseline intensity is augmented by the influence of past events. For mathematical tractability and interpretability, such models usually adopt simple parametric kernels (e.g., exponential decay), which also assume stationarity—that is, the influence depends only on the difference between the time and location of past events and those under consideration. However, real-world data often exhibit more complex, nonstationary dependencies that violate these basic assumptions. Recent work has integrated deep learning architectures into point process models, leveraging the representational power of neural networks to capture intricate event patterns beyond what simple parametric forms allow. These developments have led to two principal methodological streams: modeling the conditional intensity function \citep{du2016recurrent, mei2017neural, omi2019fully, zuo2020transformer, zhang2020self, zhou2022neural}, and modeling the influence kernel function \citep{okawa2021dynamic, zhu2022neural, dong2023spatio}. In addition, generative modeling of discrete events \citep{yuan2023spatio, ludke2023add, dong2023conditional} represents an emerging and promising direction worthy of exploration in future research.

Despite significant advances, relatively few reviews synthesize the burgeoning literature on STPPs enhanced with deep learning. For instance, \cite{rodriguez2018learning} provides a tutorial on neural temporal point processes, but it predates many recent breakthroughs; meanwhile, the surveys by \cite{yan2019recent} and \cite{Shchur2021} focus on auto-regressive networks for temporal point processes, omitting spatio-temporal settings and alternative modeling families. A recent survey \citep{mukherjee2025neural} broadly covers deep point processes based on conditional intensity functions and deep kernel formulations, but does not delve into technical details. Meanwhile, established reviews of STPPs \citep{gonzalez2016spatio, reinhart2018review, bernabeu2024spatio} primarily emphasize statistical foundations and inference for traditional (non-neural) models. 

This article addresses the need for a comprehensive overview of deep learning–based STPPs. We begin by summarizing the fundamentals of STPPs and then discuss the family of modern deep-learning kernel-based STPP frameworks, including the model architectures, survey key results, and their pros and cons. We also examine model inference techniques, including likelihood-based and likelihood-free methods, as well as recent developments in causal discovery and uncertainty quantification. We conclude by illustrating these frameworks in practical applications—earthquake analysis, infectious disease modeling, and crime dynamics—to demonstrate their versatility in tackling complex spatio-temporal problems. Through this synthesis, we aim to consolidate progress in the field and illuminate directions for future research.


\section{Background: Classical self-exciting point process models}\label{sec:background}


We first review the classical formulation of spatio-temporal point processes (STPPs) \citep{reinhart2018review}. Let \(\{(t_i, s_i)\}_{i=1}^n\) be a sequence of \(n\) events, where each event is recorded by its occurrence time \(t_i \in [0, T]\) and location \(s_i \in \mathcal{S}\subset \mathbb{R}^2\). 

Denote the event history prior to time \(t\) by 
\[
\mathcal{H}_t 
= 
\bigl\{(t_i, s_i)\mid t_i < t\bigr\},
\]
and let \(\mathbb{N}(A)\) be the counting measure that returns the number of observed events in a subset \(A \subseteq [0,T]\times \mathcal{S}\). An STPP is fully specified by its \emph{conditional intensity function}
\[
\lambda\bigl(t, s \mid \mathcal{H}_t\bigr)
\,=\,
\lim_{\Delta t, \Delta s\to 0}
\frac{\mathbb{E}
\bigl[\mathbb{N}\bigl([t,t+\Delta t]\times B(s,\Delta s)\bigr)\,\mid\,\mathcal{H}_t\bigr]}
{\Delta t\,\lvert B(s,\Delta s)\rvert},
\]
where \(B(s,\Delta s)\) is a small ball centered at \(s\) with radius \(\Delta s\). For brevity, we often write \(\lambda(t,s)\) and omit the explicit conditioning on \(\mathcal{H}_t\). The conditional intensity function can be viewed as the instantaneous rate of future events given the past, and \(\lambda(t,s)\ge 0\) must hold. 

A special case of STPPs is the \emph{self-exciting point process}, which models how past events raise (or “excite”) the likelihood of future events. Among the most prominent examples is the Hawkes process \citep{hawkes1971spectra}, defined in its simplest (temporal) form as
\begin{equation}
    \lambda(t) 
    \;=\; 
    \mu(t) 
    \;+\; 
    \sum_{t_j < t}\phi\bigl(t - t_j\bigr),
    \label{eq:pp-with-influence-kernel}
\end{equation}
where \(\mu(\cdot)\) is a deterministic baseline rate and \(\phi(\cdot)\ge 0\) is an \emph{influence function} (e.g., an exponential decay). The Hawkes process can be extended to a multi-dimensional setting, as done in the original paper  \citep{hawkes1971spectra}.
The self-exciting point process can also be extended to the spatiotemporal setting: the most commonly used model is the Epidemic-Type Aftershock Sequence (ETAS) model \citep{ogata1988statistical, ogata1998space}, widely applied to capture seismic activity, where the influence kernel \(k(\cdot)\) is defined both over time and space:
\begin{align}
    \lambda\bigl(t, s\bigr) 
    &= \mu\bigl(t, s\bigr) 
    \;+\; 
    \int_0^t \!\!\int_{\mathcal{S}} \!k\bigl(t, t', s, s'\bigr)
    \,d\mathbb{N}\bigl(t', s'\bigr) 
    \nonumber\\
    &= \mu\bigl(t, s\bigr) 
    \;+\;
    \sum_{t_j < t}\,
    k\bigl(t, t_j, s, s_j\bigr).
    \label{eq:general-form}
\end{align}
ETAS employs parametric kernels such as the Gaussian kernel, which are often assumed to be \emph{stationary}, depending only on \(\bigl(t - t'\bigr)\) and \(\bigl(s - s'\bigr)\). While this assumption simplifies computation, it may be too restrictive for modern, complex datasets.

In summary, most classic self-exciting STPP models rely on parametric, stationary kernels to ensure tractability and interpretability. However, many real-world applications involve heterogeneous patterns that do not conform to strict stationarity or simple exponential decay. This motivates more flexible approaches that allow for non-stationary, potentially high-dimensional influence functions, which we review in the subsequent sections.

\section{General deep influence kernel for spatio-temporal process}

Real-world spatio-temporal data often exhibit non-stationary and non-homogeneous dynamics, where the magnitude or shape of event-triggering effects changes over time or depends intricately on location. Traditional exponential-decay kernels are limited in their ability to capture complex phenomena such as sudden bursts, long-tail decays, or structural inhomogeneities in space. To address these limitations, recent work has integrated deep learning architectures into point process models, leveraging the expressive power of neural networks to learn complex event patterns that go beyond the capabilities of simple parametric forms. In this section, we review these developments.

\subsection{Main approaches}\label{sec:main_approaches}

Modern approaches to generalizing self-exciting Hawkes processes can be broadly divided into two categories. One category focuses on representing the conditional intensity function with neural architectures, such as recurrent neural networks (RNNs) \citep{du2016recurrent, xiao2017modeling, chen2020neural, zhu2021imitation}, attention-based layers \citep{zuo2020transformer, zhang2020self}, or other advanced sequence models. While these methods often capture complex dependencies in event sequences, they treat the model largely as a “black box,” thereby obscuring how individual past events shape future intensities. Furthermore, most of these neural intensity models were designed for one-dimensional or purely temporal data, limiting direct application to continuous spatio-temporal settings.

The other category retains a Hawkes-style additive structure while generalizing the kernel itself. Instead of adopting exponential or Gaussian decay, the kernel is represented via neural networks or other flexible parameterizations. Such methods preserve the explicit notion of how each past event exerts an influence on future events, allowing the kernel to adapt to more intricate patterns than standard parametric forms. Representative examples include dynamic kernels for time intervals \citep{okawa2021dynamic}, neural spectral kernels \citep{zhu2022ICLRneural}, and other deep learning approaches to kernel modeling \citep{zhu2022neural}. These works demonstrate that moving beyond a fixed decay function can substantially improve a model’s ability to capture nonstationary and high-dimensional dependencies, including continuous space-time or graph-based data. 

\begin{figure}[!htb]
\centering
  \includegraphics[width=.9\textwidth]{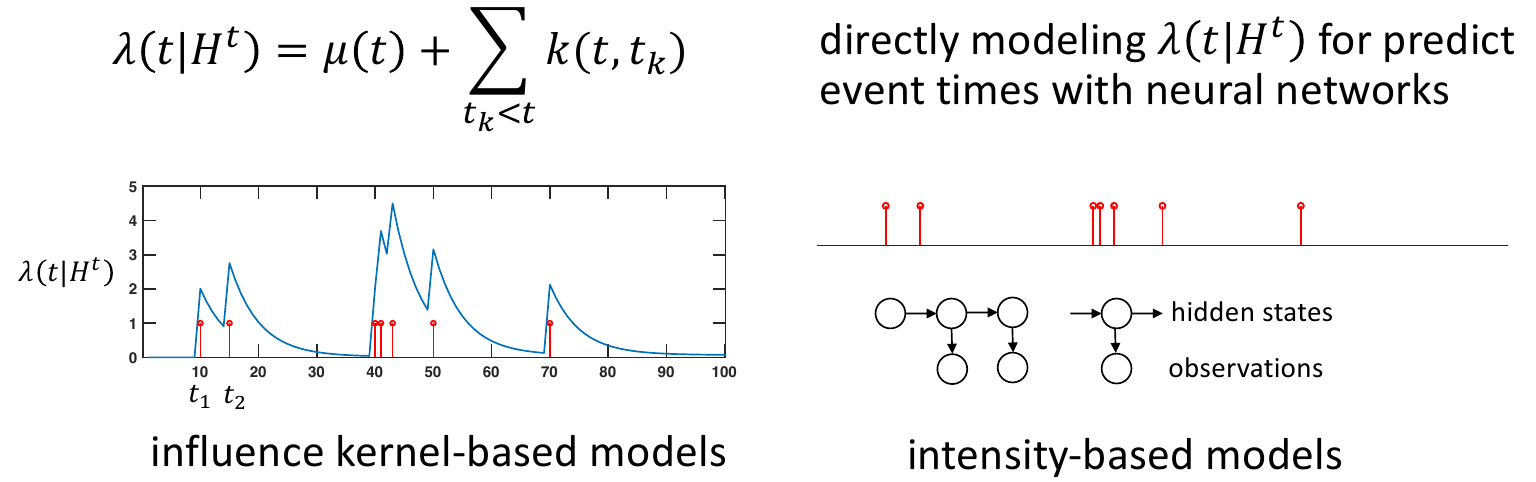} 
 \vspace{.1in}
 \caption{Two types of deep learning-based models for self-exciting point processes, \textcolor{black}{illustrated for one-dimensional temporal case here.  On the left panel, for ``influence kernel–based models,'' the red bars denote individual events and the blue line represents the conditional intensity function. Each event causes an increase in the intensity, and the effects of past events are additive, leading to clustered future events. On the right panel, illustrating ``intensity-based models,'' there is no direct parametric specification of the conditional intensity function; instead, it is modeled through a recursive neural network–type architecture.}
 }
 \label{fig:models}
 \end{figure}

Following this perspective, we can categorize modern spatio-temporal point process (STPP) models into two main types, as illustrated in {\bf Figure}~\ref{fig:models}:
\begin{itemize}

\item Direct modeling of the conditional intensity: 
   Here, the intensity \(\lambda(t,s)\) is parameterized via autoregressive neural architectures like RNNs \citep{du2016recurrent} or continuous-time LSTMs \citep{mei2017neural}, and later by self-attention mechanisms \citep{zhang2020self, zuo2020transformer}. Although this approach can achieve strong predictive accuracy, interpretability often remains limited since the influence of a past event is not explicitly traced through a kernel function.

\item Kernel-based modeling of self-excitation: 
   In this framework, one maintains the additive summation of influences over past events, but the kernel function \(k(\cdot)\) is learned flexibly—e.g., with deep neural networks \citep{zhu2022ICLRneural, chen2020neural} or dynamic kernel forms \citep{okawa2021dynamic}. By doing so, the model preserves an interpretable self-exciting structure yet allows more complex functional forms than standard exponential or Gaussian assumptions.
  
 \end{itemize}
Table~\ref{tab:literature-review} provides a more complete summary of these approaches in terms of their model formulations and key features.


In this article, we focus on the kernel-based modeling paradigm, as it offers an interpretable framework for capturing how past events influence future occurrences, thereby facilitating an understanding of complex spatio-temporal dynamics. Unlike direct intensity-based models, which often obscure the contribution of individual historical events, kernel-based approaches make the structure of temporal, spatial, or relational dependencies explicit. This interpretability is particularly valuable in applications such as earthquake aftershock modeling, epidemic spread, and urban crime analysis, where identifying localized triggers and propagation patterns is critical. To further enhance the expressiveness of kernel models, we emphasize the use of deep kernels: low-rank representations that relax traditional stationarity assumptions and allow for variation across time, space, or graphs.

\begin{table}[!h]
  \caption{Comparison between common parametric and deep learning-based methods for self-exciting point process modeling. \textcolor{black}{We compare the methods along several dimensions: whether they are parametric or non-parametric, whether the conditional intensity function is modeled using an influence kernel, whether the method supports spatio-temporal data, and whether Graph Neural Networks (GNNs) are incorporated.  The table is intended as a quick reference, where readers can compare modeling assumptions and structural features at a glance.}}
  
  \vspace{.1in}
  \centering

  \begin{threeparttable}

  \resizebox{\linewidth}{!}{%
  \begin{tabular}{lccccccccc}
    \toprule
    & RMTPP & NH & FullyNN & SAHP & LogNormMix & DeepSTPP & NSTPP & DSTPP & DNSK \\
    \midrule
    Non-parametric            & \cmark & \cmark & \cmark & \cmark & \cmark & \xmark & \cmark & \cmark & \cmark \\
    Modeling influence kernel & \xmark & \xmark & \xmark & \xmark & \xmark & \xmark & \xmark & \xmark & \cmark \\
    Spatio-temporal           & \xmark & \xmark & \xmark & \xmark & \xmark & \cmark & \cmark & \cmark & \cmark \\
    Using GNN                 & \xmark & \xmark & \xmark & \xmark & \xmark & \xmark & \xmark & \xmark & \xmark \\
    \bottomrule
  \end{tabular}%
  }

  \vspace{0.6em}

  \resizebox{\linewidth}{!}{%
  \begin{tabular}{lcccccccc}
    \toprule
    & GNHP & GINPP & THP\text{-}S & SAHP\text{-}G & GHP & GNPP & GBTPP & GraDK \\
    \midrule
    Non-parametric            & \xmark & \cmark & \cmark & \cmark & \xmark & \cmark & \xmark & \cmark \\
    Modeling influence kernel & \xmark & \cmark & \xmark & \xmark & \cmark & \xmark & \xmark & \cmark \\
    Spatio-temporal           & \xmark & \xmark & \xmark & \xmark & \xmark & \xmark & \xmark & \cmark \\
    Using GNN                 & \xmark & \cmark & \xmark & \xmark & \cmark & \cmark & \cmark & \cmark \\
    \bottomrule
  \end{tabular}%
  }

  \begin{tablenotes}[flushleft]
    \footnotesize
    \item \textbf{Model abbreviations:} RMTPP – \citet{du2016recurrent}, NH – \citet{mei2017neural}, FullyNN – \citet{omi2019fully}, SAHP – \citet{zhang2020self}, LogNormMix – \citet{shchur2019intensity}, DeepSTPP – \citet{zhou2022neural}, NSTPP – \citet{chen2020neural}, DSTPP – \citet{yuan2023spatio}, DNSK – \citet{dong2023spatio}, GNHP – \citet{fang2023group}, GINPP – \citet{pan2023graphinformed}, THP\text{-}S – \citet{zuo2020transformer}, SAHP\text{-}G – \citet{zhang2021learning}, GHP – \citet{shang2019geometric}, GNPP – \citet{xia2022graph}, GBTPP – \citet{wu2020modeling}, GraDK – \citet{dong2023deep}.
  \end{tablenotes}

  \end{threeparttable}

  \label{tab:literature-review}
\end{table}

\subsection{General kernel and representation}

Given the limitations of parametric kernels discussed in Section~\ref{sec:background}, we are motivated to consider a more general self-exciting point process with a flexible influence kernel.  
Specifically, let an event be denoted as \(x \in \mathcal{X} \subseteq \mathbb{R}^d\). For instance, in a temporal point process, \(x = t\) contains only the event time, so \(d = 1\). For spatiotemporal processes, \(x = (t, s)\), where \(s\) typically represents the two-dimensional spatial coordinates of the event, making \(d = 3\). In marked spatiotemporal point processes, \(x = (t, s, m)\), where, in addition to time and location, the mark \(m\) provides further information about the event—such as earthquake magnitude, crime category, or contextual information (e.g., a feature vector). Given this notation, we denote the observations as \(x_1, x_2, \ldots\); for example, in a spatiotemporal point process, each \(x_i = (t_i, s_i)\), which can be ordered by their associated event times \(t_1 < t_2 < \cdots\). Below, we present the model for spatio-temporal point processes. 
Then the conditional intensity is modeled by
\[
    \lambda(x)
    \;=\;
    \mu(x)
    \;+\;
    \sum_{j: t_j < t}
    k\bigl(x, x_j\bigr),
\]
where \(\mu(x) > 0\) is a time- and location-dependent baseline term, and \(k(x, x') : \mathcal{X} \times \mathcal{X} \to \mathbb{R}\) is a kernel that captures the influence of a past event at \(x' = (t', s')\)  on a future event at \(x = (t, s)\), with $t'<t$. This framework generalizes several well-known models: in a one-dimensional Hawkes process, \(x=t\) and \(k(x, x')=\phi(t - t')\), $t'<t$; in an ETAS model, \(k(x, x')\) reflects space-time interactions in \(x=(t, s)\)-coordinates.

\begin{textbox}\section{Mercer's theorem \citep{mercer1909functions}}
If \( K(x, x'): \mathcal X \times \mathcal X \rightarrow \mathbb \mathbb R^+ \) is a continuous, symmetric, and positive semi-definite kernel on a compact domain \( \mathcal X \), then it can be written as:
\begin{equation}
k(x, x') = \sum_{r=1}^\infty \nu_r \phi_r(x) \phi_r(x'),
\label{eqn:kernel1}
\end{equation}
where \( \{ \nu_r\geq 0 \} \) are eigenvalues, \( \{ \phi_r \} \) are the corresponding orthonormal eigenfunctions, and the series converges absolutely and uniformly on \( \mathcal X \times \mathcal X \).
\end{textbox}

A general kernel can accommodate phenomena not captured by simple parametric forms, including potential negative influence (\(k(x, x') < 0\)) and asymmetric causality (\(k(x, x') \neq k(x', x)\)): for instance, in traffic networks, upstream incidents may affect downstream conditions but not vice versa, leading to non-reciprocal interactions \citep{zhu2021spatio}.

 Mercer's Theorem can be extended to asymmetric and indefinite kernels \citep{seely1919non, jeong2024extending}. Such extensions provide a theoretical basis for us to decompose the kernel \(k\) in our setting via a finite-rank representation:
\begin{equation}
    k(x, x') 
    \;=\;
    \sum_{r=1}^R
    \nu_r \, \psi_r(x) \, \phi_r(x'),
    \quad 
    \nu_r \ge 0,
    \label{kernel}
\end{equation}
where \(\{\psi_r(\cdot)\}\) and \(\{\phi_r(\cdot)\}\) are eigenfunctions, which can also be viewed as feature maps, and \(\nu_r\) are nonnegative eigenvalues. Although infinite-dimensional expansions exist in theory, practical models often truncate to a finite rank \(R = 1, 2, \ldots\). Moreover, in practice, we find that \(\psi_r\) and \(\phi_r\) do not need to be orthogonal to achieve good empirical performance, and \(k\) can be indefinite or asymmetric as long as they violate the fundamental requirement \(\lambda(x)\ge 0\).

\textcolor{black}{Such a kernel representation (through eigenfunctions) allows for non-stationarity in space and time, since the influence function is not restricted to depend solely on the difference between event times, $t - t'$, and/or the difference between event locations, $s - s'$. The kernel representation is based on Mercer's Theorem (or its extensions) in the most general case.}
The kernel decomposition in Equation \eqref{kernel}, by adopting flexible eigenfunctions, can represent diverse non-stationary phenomena by allowing nonlinear time and spatial dependencies. However, such flexibility introduces considerable challenges in parameter estimation and model identifiability, which must be addressed to ensure reliable inference and tractable computation.

\subsection{Constructing low-rank kernel}\label{low-rank-kernel}

A key insight is that one can obtain a low-rank approximation of the influence kernel. In this section, we focus on the formulation for spatio-temporal kernels (with temporal kernels viewed as a special case). The construction of influence kernels for point processes defined on graphs is discussed in Section~\ref{sec:GNN}. \textcolor{black}{This approach has been utilized in constructing spatio-temporal kernel in \cite{dong2023spatio}.}

For a non-stationary spatio-temporal kernel, we reparameterize it in terms of temporal and spatial displacements. \textcolor{black}{In essence, this reparameterization makes the kernel easier to approximate, automatically handling both stationary and nonstationary behaviors.} For instance, in the spatiotemporal setting, instead of viewing the influence kernel \(k(x', x)\) with coordinates \((t', s')\) and \((t, s)\) as \(k(t', t, s', s)\), we rewrite it as  
\[
    k\bigl(t',\, t - t',\, s',\, s - s'\bigr),
\]  
where \(s - s'\) denotes element-wise difference when the spatial domain is multi-dimensional. This reparameterization preserves the original information but introduces a structure that is more amenable to low-rank representation. Following the theory of kernel decomposition \citep{mercer1909functions, mollenhauer2020singular}, we arrive at 
\begin{equation}
  k\bigl(t',\,t - t',\,s',\,s - s'\bigr) 
  \;=\; 
  \sum_{r=1}^{R}\sum_{l=1}^{L} 
  \alpha_{lr}\,\psi_l\!\bigl(t'\bigr)\,u_r\!\bigl(s'\bigr)\,\varphi_l\!\bigl(t - t'\bigr)\,v_r\!\bigl(s - s'\bigr),
  \quad
  t>t'.
  \label{eq:spatio-temporal-kernel-representation}
\end{equation}
Here, \(\{\psi_l,\varphi_l\}\) capture how an event at time \(t'\) influences future times \(t\), and \(\{u_r,v_r\}\) encode initial and propagated spatial effects. The weights \(\alpha_{lr}\) then combine these basis functions in a flexible summation, leading to a flexible kernel approximation.
Note that if \(\psi_l(t')\) and \(u_r(s')\) remain constant, the kernel reverts to the form of a stationary kernel.  Hence, the representation automatically accommodates both stationary and nonstationary behaviors. This kernel decomposition has been used in \cite{dong2023spatio}, called the \texttt{DNSK}.

In practice, one must determine the kernel rank to be used in modeling the data, and there are two possible approaches. The first approach treats the kernel rank as a hyperparameter and tunes it via cross-validation. The second approach treats the kernel rank as a level of model complexity to be learned directly from the data. Under regularity assumptions for kernel decomposition, the singular values decay to zero, yielding a low-rank approximation. These singular values appear in the coefficients \(\alpha_{rl}\). The effective rank is then determined by retaining only the coefficients of significant magnitude, with no need to pre-specify the rank.

{\it Example: Low-rank representation for a temporal-only process.} To illustrate and develop intuition, we present an example of a temporal kernel with time discretization.  
In this example, we demonstrate that the construction indeed leads to a lower-rank representation of the kernel. Consider the following synthetic kernel:  
\[
k(t, t') = 0.3 \sin(1.2 t')\,\sin(2 (t - t'))\,
e^{-0.5(t - t')}/(1 + e^{5(t - t' - 3)}), \quad \text{for } t > t'.
\]  
The kernel \( k(t, t') \), as well as its reparameterized form \( k(t', t - t') \), is evaluated on a discrete time grid \( t, t' = 1, \ldots, 200 \), with \( t' < t \). One can observe a drastic difference in rank: the original parameterization has a rank of 197, whereas the reparameterization has a rank of 1. Note that, due to temporal causality, part of the matrix is unspecified (i.e., \( k(t, t') \) is only defined for \( t' < t \)). If the unspecified entries are appropriately filled, the rank can be further reduced; the current result is based on zero-filling the unknown entries.

\begin{textbox}\section{Deep kernel: Main idea}

The main idea of the deep kernel point process is to represent the influence kernel using neural networks. Drawing ideas from Mercer's decomposition, the basis functions are parameterized by neural architectures without the need for orthogonality or normalization. This design exploits the expressive capacity of neural networks and allows the kernel to capture both positive and negative influences. Nonnegativity of the resulting intensity function is then enforced through suitable constraints in the model-learning optimization process.
\end{textbox}

\subsection{Neural network-based deep-kernel}

A sensible and popular approach for modeling kernel functions is to represent their eigenfunctions \(\psi_r\) and \(\phi_r\) using neural networks; \textcolor{black}{many prior works have taken such an approach, for instance, \cite{zhu2021imitation,zhu2022neural,dong2023spatio,dong2023deep}.} By leveraging the universal approximation capabilities of deep learning, this approach can capture complex spatio-temporal dependencies and approximate a broad range of kernel forms. Moreover, adopting a low-rank decomposition confines learning to a finite set of temporal and spatial components, with higher-order modes truncated to maintain computational efficiency. 

In practice, many methods use fully connected neural networks to construct these basis functions. Inputs (e.g., historical event embeddings or spatial displacements) are mapped into a hidden feature space through a shared multi-layer sub-network, commonly with Softplus activations. The resulting embeddings are passed to \(R\) distinct sub-networks, each producing one of the spatial basis functions \(\{\phi_r(x)\}_{r=1}^R\). Consequently, a kernel with temporal rank \(L\) and spatial rank \(R\) may involve \(2(L + R)\) sub-networks in total—one set for temporal components and another for spatial components. To accommodate inhibitory effects, some implementations use a linear (unbounded) output layer in the temporal sub-networks, thereby allowing negative values in the kernel.

As mentioned earlier, there is no strict requirement for orthogonality among the basis functions; empirical findings suggest that relaxing this constraint does not harm performance and may even enhance flexibility. Although multilayer perceptrons with Softplus or sigmoid activations are commonly used, alternative architectures can be equally effective. For instance, when data lie on structured domains such as graphs, graph neural networks (GNNs) might be more suitable, and spline-based models or other function approximators can also be employed. 


\subsection{Marked point process and high-dimensional marks}\label{sec:high-dim-mark}

\textcolor{black}{In principle, } the framework can be extended to model marked STPPs. \textcolor{black}{Similar approach has been taken by \cite{zhu2022spatiotemporaltextual}, although such construction in practical setup remains under explored.} In particular, we can introduce additional sets of mark basis functions \(\{g_q, h_q\}_{q=1}^{Q}\). In this case, the influence kernel function \(k\) becomes, for $t'<t$,
\[
k\bigl(t^\prime,\, t - t^\prime,\, s^\prime,\, s - s^\prime,\, m^\prime,\, m\bigr) 
= 
\sum_{q=1}^{Q} 
\sum_{r=1}^{R} 
\sum_{l=1}^{L} 
\alpha_{lrq}\,\psi_l(t^\prime)\,\varphi_l(t - t^\prime)\,u_r(s^\prime)\,v_r(s - s^\prime)\,g_q(m^\prime)\,h_q(m),
\]
where \(m^\prime, m \in \mathcal{M} \subset \mathbb{R}^d\), where $d$ is the dimension of the mark space, and \(\{g_q, h_q\}\) can be represented by neural networks that capture the influences of the historical mark \(m^\prime\) and current mark \(m\), and can also be viewed as feature maps. Note that, in many applications, the mark space \(\mathcal{M}\)  can be complex and discrete (such as categorical variables), learning separate functions \(g_q\) and \(h_q\) is more appropriate than modeling \(m-m^\prime\).  We will provide one real-data example to demonstrate the use of deep kernel for high-dimensional marks in Section \ref{sec:high-dim-mark}.

\subsection{Kernel based on graph neural networks (GNN)}\label{sec:GNN}

In a point process defined over a graph, we observe multi-dimensional point processes at each of the graph’s nodes. The underlying graph structure determines how events occurring at different nodes may influence one another. The graph-based model can also be related to spatial models; for instance, by discretizing a continuous space, one can treat each location as a node in the graph. This allows for more flexible “non-Euclidean” modeling, where nodes corresponding to spatially distant locations may still be directly connected. 

Consider a graph with vertices \(v \in V\). The underlying graph topology may be given {\it a priori} or learned directly from data. In the graph point process setting, each observation takes the form \((t_i, v_i)\), where \(v_i \in V\) specifies the node where the $i$-th event occurs. Let \(\{(t_i, v_i)\}_{i=1}^{n}\) denote the sequence of \(n\) observed events.
The conditional intensity \(\lambda(t,v)\), given an influence kernel \(k\), is defined as
\begin{equation}
    \lambda(t, v) = \mu \;+\; \sum_{j: t_j < t} \; k(t, t_j, v, v_j).
    \label{eq:conditional-intensity-with-kernel}
\end{equation}
Note that the node index \(v\) for the observation can be treated as a discrete mark, and thus, the process can be viewed as a marked point process. Moreover, one may also observe additional marks associated with the node. In such cases, the mark \(v\) may represent continuous or categorical characteristics of the event, such as location or type.

A common approach to specifying \(k\) in the graph-based point process setting \citep{reinhart2018review} is given by
\[
    k(t', t, v', v) = a_{v, v'} \, f(t - t'),
\]
where \(a_{v, v'}\) measures the influence of node \(v'\) on \(v\) through a graph-based kernel, and \(f\) is a stationary temporal kernel; it is commonly assumed $a_{v, v'}\geq 0$. In this formulation, the influence over time and over the graph is decoupled; that is, only nodes with direct edges exert influence on each other.


 In contrast, we will present a GNN-based kernel that enables a more expressive representation of graph-based influence, allowing the model to capture richer event dynamics. Extending the influence kernel to operate on a graph structure while incorporating graph neural network (GNN) architectures is both essential and non-trivial for modeling point processes on graphs. The challenges arise because: (i) the usual notion of distance between event marks is not directly applicable in a graph setting, rendering distance-based spatial kernels ineffective, and (ii) replacing distance-based kernels with scalar coefficients to capture pairwise node interactions significantly restricts model expressiveness in modern applications. 

We employ the basis kernels expansion strategy similar to before. In particular, to model the graph dependency, we handle by graph filter design (see, e.g., \cite{dong2023deep}). Specifically, the influence kernel for influence across time and nodes, in Equation \eqref{eq:conditional-intensity-with-kernel} is decomposed into basis kernel functions as follows:
\begin{equation}
    k(t, t', v, v') = \sum_{r=1}^{R}\sum_{l=1}^{L}\alpha_{rl}\psi_l(t^\prime)\varphi_l(t-t^\prime)B_r(v^\prime, v),
    \label{eq:temporal-graph-decomposed-kernel}
\end{equation}
where $\alpha_{rl}$ are the coefficients, $\{\psi_l, \varphi_l\}_{l=1}^{L}$ are sets of eigenfunctions for the (possibly nonstationary) temporal kernel, and $B_r: V\times V\rightarrow \mathbb R$, $r = 1, \ldots, R$ are the graph filters.  Here we``separately'' model event dependency over time or graph using different basis kernels. The temporal is expanded using the strategy in Section \ref{low-rank-kernel}, and the graph kernel $h_r(v', v)$ can be implemented using specific graph neural networks (GNN), as described in the following. 

GNN process signals on graphs by cascading linear operations with pointwise nonlinearities that incorporate the underlying graph structure \citep{bruna2013spectral,defferrard2016convolutional,wu2020comprehensive}.  In general, each layer of a GNN takes the form \(\sigma(\Theta x)\), where \(x \in \mathbb{R}^d\) is the input signal, \(\Theta \in \mathbb{R}^{d \times d}\) is a learnable weighting matrix, and \(\sigma(\cdot)\) is a pointwise nonlinear function. In the specific case of graph convolutional neural networks, \(\Theta\) is defined by a localized graph filter that encodes the graph topology \citep{bruna2013spectral,defferrard2016convolutional}.  

To motivate this construction, we first clarify the notion of convolution on a graph. Let \(G = (V, E)\) be a graph with node set \(V\) and edge set \(E\). For a graph with \(N\) nodes, let \(A, D \in \mathbb{R}^{N \times N}\) denote the adjacency matrix and the degree matrix, respectively, and let \(L = D - A\) be the graph Laplacian. Classical convolution from spatial or temporal signal processing extends to the graph domain through the concept of \textit{graph filters} \citep{ortega2018graph,leus2023graph}, which can be formulated from two perspectives: spectral and algebraic. From the \emph{spectral} viewpoint \citep{hammond2011wavelets,shuman2013emerging}, one defines graph filters via the eigen-decomposition of \(L\), paralleling the role of Fourier transforms in standard signal processing. From the \emph{algebraic} viewpoint \citep{sandryhaila2013discrete,sandryhaila2014discrete}, one treats the adjacency matrix \(A\) as a shift operator and constructs graph filters as matrix polynomials in \(A\). These two perspectives unify under the general framework:
\[
x \ast B = Bx,
\]
where $\ast$ denotes convolution, \(B \in \mathbb{R}^{N \times N}\) is a \textit{localized graph filter} \citep{leus2023graph} and \(x \in \mathbb{R}^N\) is a signal on the graph. Concretely, one may write
\[
x \ast B = U \, \mathrm{diag}(f)\, U^{-1} x 
\quad\text{or}\quad
x \ast B = \sum_{j=1}^J h_j \, S^j x,
\]
where \(S = U \Lambda U^{-1} \in \mathbb{R}^{N \times N}\) (e.g., \(S = A\) or \(S = L\)) encodes the graph structure, \(f \in \mathbb{R}^N\) is the frequency-domain representation of the filter, and \(\{h_j\}\) are polynomial coefficients. In this work, we use \(B(v_i, v_j)\) to denote \(B_{ij}\), which represents the influence of node \(v_i\) on node \(v_j\) under the localized filter \(B\). Based on this, the graph filters will be represented via graph-based kernels by leveraging the localized graph filters in graph convolution to extract informative inter-event-category patterns from graph-structured data. We will provide an example to demonstrate the use of deep kernel using non-stationary temporal kernel and GNN for capturing graph structures in Section \ref{sec:GNN_dp}.

\subsection{Discrete space, discrete time: Connection to non-linear time-series}

One approach to the kernel recovery problem is to discretize both space and time, resulting in a fully discrete model formulation (see, e.g., \cite{juditsky2020convex,juditsky2023generalized}). In this case, the problem reduces to a nonlinear time series model. For instance, when events occur sparsely, the observations can be represented as binary variables \(\omega_{jk} \in \{0,1\}\), where \(j = 1, 2, \ldots\) indexes discrete time and \(k = 1, \ldots, K\) indexes spatial locations. The conditional probability of an event occurring at time \(j\) and location \(k\) given the past history \(\mathcal{H}_t\) can be modeled as  
\begin{equation}
\mathbb{P}\{\omega_{jk}=1 \mid \mathcal{H}_t\} = \beta_0 + \sum_{\ell=1}^K \sum_{1 \leq i \leq d} \beta_{k\ell i} \omega_{j-i,\ell}, \label{bern_model}
\end{equation}
where \(\theta = (\beta_0, \beta_{k\ell i}; 1 \leq k, \ell \leq K, 1 \leq i \leq d)\) denotes the model parameters, which can be viewed as a discretized representation of kernels, and \(d\) represents the memory depth, indicating how far into the past history the model looks. The feasible domain of $\theta$ is such that the parameter leads to a conditional probability in Equation \eqref{bern_model} to be in the range [0, 1]. This discrete formulation connects naturally to nonlinear time series models, where the coefficients \(\beta_{k\ell i}\) can be interpreted in terms of Granger causality since the coefficients $\beta_{k\ell i}$ can be used to infer Granger causality (see, e.g., \cite{shojaie2022granger}). Furthermore, the estimation of such models can be approached using techniques based on monotone variational inequalities for estimating the generalized linear models (GLM) (see, e.g., \cite{juditsky2020convex}).

\section{Model estimation}

We now discuss \emph{kernel estimation} for STPPs, where the aim is to infer the influence kernel \(K(x, x')\) from an observed event history \(\mathcal{H}_t = \{(t_i, s_i)\,:\, t_i < t\}\). This kernel governs how past events influence future dynamics, making the estimation task analogous to solving an inverse problem: given discrete event data, one seeks to recover the continuous influence function. Although reminiscent of system identification, our setting is unique in that it treats both space and time as continuous domains.

We examine several common kernel estimation approaches for STPPs, including Maximum Likelihood Estimate (MLE) and Least-Squares (LS) recovery. The computational complexity associated with STPP estimation has long been acknowledged \citep{veen2008estimation, schoenberg2018comment}, and it becomes particularly demanding when neural network structures are introduced. Recent studies \citep{dong2023spatio, libeyond, yuan2023spatio} propose various approaches to address the computational burden in estimating neural STPPs. For example, \citet{dong2023spatio} introduces a log-barrier method for the maximum likelihood estimation (MLE) problem, reducing complexity from \(\mathcal{O}(n^3)\) to \(\mathcal{O}(n)\). Other works favor likelihood-free strategies, such as minimizing the Wasserstein distance between event distributions \citep{xiao2017modeling}, using score-based methods \citep{libeyond}, or employing advanced generative models to represent discrete events directly \citep{yuan2023spatio, ludke2023add}. Nonetheless, the ongoing challenge of improving STPP estimation efficiency continues to motivate the exploration of new approaches. 

\textcolor{black}{In this section, we focus on computational strategies for the deep kernel decomposition introduced earlier, with an emphasis on reducing computational complexity through pre-computation and approximation. A notable advantage of representing the conditional intensity function using influence kernels is that portions of the likelihood or loss function can be pre-computed, a benefit also shared by classical parametric kernels (see, e.g., \cite{reinhart2018review}). Related work has also investigated approximating the likelihood computation for conditional intensity functions represented by neural networks, such as \cite{zhou2023automatic}. While our emphasis is on deep kernels, many of these ideas—such as pre-computation and approximation, may inspire analogous strategies for other models in Section \ref{sec:main_approaches}.}

\subsection{Maximum Likelihood Estimate (MLE) }\label{sec:MLE}

Given the conditional intensity function in Equation \eqref{eq:pp-with-influence-kernel}, the log-likelihood for an observed event sequence \(\mathcal{H}_T\) is
\begin{equation}
    \ell(\theta)
    \;=\;
    \sum_{i=1}^n \log \lambda\bigl(t_i, s_i\bigr)
    \;-\;
    \int_0^T \!\int_{\mathcal{S}} \lambda(t, s)\,ds\,dt,
    \label{eq:pp-log-likelihood}
\end{equation}
where \(\theta\) encompasses all model parameters. Hawkes process models can then be estimated by solving \(\max_{\theta}\ell(\theta)\). In practice, two standard approaches for solving this MLE problem are the EM algorithm (also known as stochastic declustering) and gradient-based methods \citep{reinhart2018review}. 

Earlier studies \citep{veen2008estimation, reinhart2018review} demonstrate that EM  can achieve analytic iterations with a closed-form expression to maximize the likelihood by introducing a latent variable \(u_i\) for each event. The EM algorithm is numerically stable when the number of events is not too large; however, it is not scalable for large numbers of events, as it requires introducing a set of auxiliary variables for each event, which scales quadratically in \(n\). Thus, for modern neural STPPs with deep networks, gradient-based methods solving MLE are more amenable to large-data settings. 

In computing the log-likelihood objective function in Equation \eqref{eq:pp-log-likelihood}, many early neural STPP approaches rely on direct numerical integration and summation to compute this log-likelihood. In Equation \eqref{eq:pp-log-likelihood}, the summation term \(\sum_i \log \lambda(t_i, s_i)\) is relatively straightforward: one need only compute \(\lambda\) at the observed events, which involves calculating kernel contributions \(\alpha_{lr}\,\psi_l(t_j)\,\varphi_l(t_i - t_j)\,u_r(s_j)\,v_r(s_i - s_j)\) for any \(t_j < t_i\). To contain costs, many implementations assume a limited region of influence such that events with \(\lvert t_i - t_j\rvert\) and \(\|s_i - s_j\|\) exceeding certain thresholds do not contribute. This truncation makes the computation scalable in large datasets.

\subsubsection{Computation of integral} \label{sec:int} 

The most computationally intensive component in evaluating the log-likelihood in \eqref{eq:pp-log-likelihood} for general STPPs with nonparametric influence functions typically arises from the required numerical integration. One approach to circumvent this is by decomposing the integral into integrals over basis functions, leveraging the aforementioned low-rank basis decomposition. In particular, for spatio-temporal kernel in Equation \eqref{eq:spatio-temporal-kernel-representation}, we have
\begin{align}
    \int_{0}^{T}\!\int_{\mathcal{S}}\!\lambda(t, s)\,ds\,dt 
    &= 
    \mu \lvert \mathcal{S}\rvert T 
    \;+\; 
    \sum_{i=1}^{n}\int_{0}^{T}\!\int_{\mathcal{S}}\!
    \mathbb{I}(t_i < t)\,
    k\bigl(t_i,\,t,\,s_i,\,s\bigr)\,ds\,dt 
    \nonumber\\[5pt]
    &= 
    \mu \lvert \mathcal{S}\rvert T 
    \;+\;
    \sum_{i=1}^{n}
    \sum_{r=1}^{R} 
    u_r\bigl(s_i\bigr)
    \Bigl(\!\int_{\mathcal{S}}v_r\!\bigl(s - s_i\bigr)\,ds\Bigr)
    \sum_{l=1}^{L}\!
    \alpha_{rl}\,\psi_l\bigl(t_i\bigr)
    \int_{0}^{T-t_i}\!\varphi_l(t)\,dt,
    \label{eq:double-integral}
\end{align}
where \(\{\varphi_l\}\) and \(\{v_r\}\) are evaluated on dense grids. 
When the effective range of influence is finite (i.e., beyond certain \(\tau_{\text{max}}\) and \(a_{\text{max}}\), influence becomes negligible), one can further restrict \(\{\varphi_l\}\) and \(\{v_r\}\) to the domain \([0, \tau_{\text{max}}]\times B(0, a_{\text{max}})\) rather than \([0,T]\times\mathcal{S}\), thereby reducing computational overhead.

A related challenge arises when the event marks are high-dimensional, as integrating over large-dimensional spaces can be prohibitively expensive. Two possible remedies include exploiting latent low-dimensional structures—for instance, using a small set of latent features despite high ambient dimensionality—or encoding/embedding methods. The latter encompasses, for example, binary encodings or learned text embeddings for police report data \citep[e.g.,][]{zhu2022spatiotemporaltextual}, both of which alleviate computational costs and allow scalable estimation of marked STPPs with complex marks. 


\subsection{Ensuring non-negativity of intensity}

Ensuring \(\lambda(t, s) \ge 0\) is essential for the conditional intensity function of a point process. Since the influence kernel may take negative values, additional care is needed to guarantee the non-negativity of the resulting intensity. This is achieved by explicitly enforcing the constraint during optimization. A common approach is to apply a scaled positive transformation to maintain \(\lambda\ge 0\), such as a nonnegative activation function, e.g., Softplus \citep{mei2017neural, zuo2020transformer, zhang2020self}. Alternatively, the barrier method—widely used in optimization \citep{kelaghan1982barrier, li2003fast}—can be employed. Below, we present a log-barrier approach to incorporate the non-negativity constraint directly into the objective function.
Let \(\ell(\theta)\) denote the log-likelihood from Equation \eqref{eq:pp-log-likelihood}, recast in terms of the model parameters \(\theta\). The constrained MLE problem can be written as
\[
    \min_{\theta} \;-\ell(\theta)
    \quad
    \text{subject to}
    \quad
    -\lambda(t,s)\,\le 0,
    \;\forall\,t\in [0, T],\, s\in \mathcal{S}.
\]
To incorporate this constraint, we apply a log-barrier function \citep{boyd2004convex} that penalizes instances where \(\lambda(t,s)\) might approach zero. The log-barrier method preserves the linear form of \(\lambda\) and promotes computational efficiency in evaluating the integration of the intensity function in computing the log-likelihood function as discussed in Section \ref{sec:int}, while enhancing the model’s ability to recover the underlying kernel and intensity.

\subsubsection{Example: Recover a low-rank spatio-temporal kernel}
\begin{figure}[!htb]
\centering

\includegraphics[width=\linewidth]{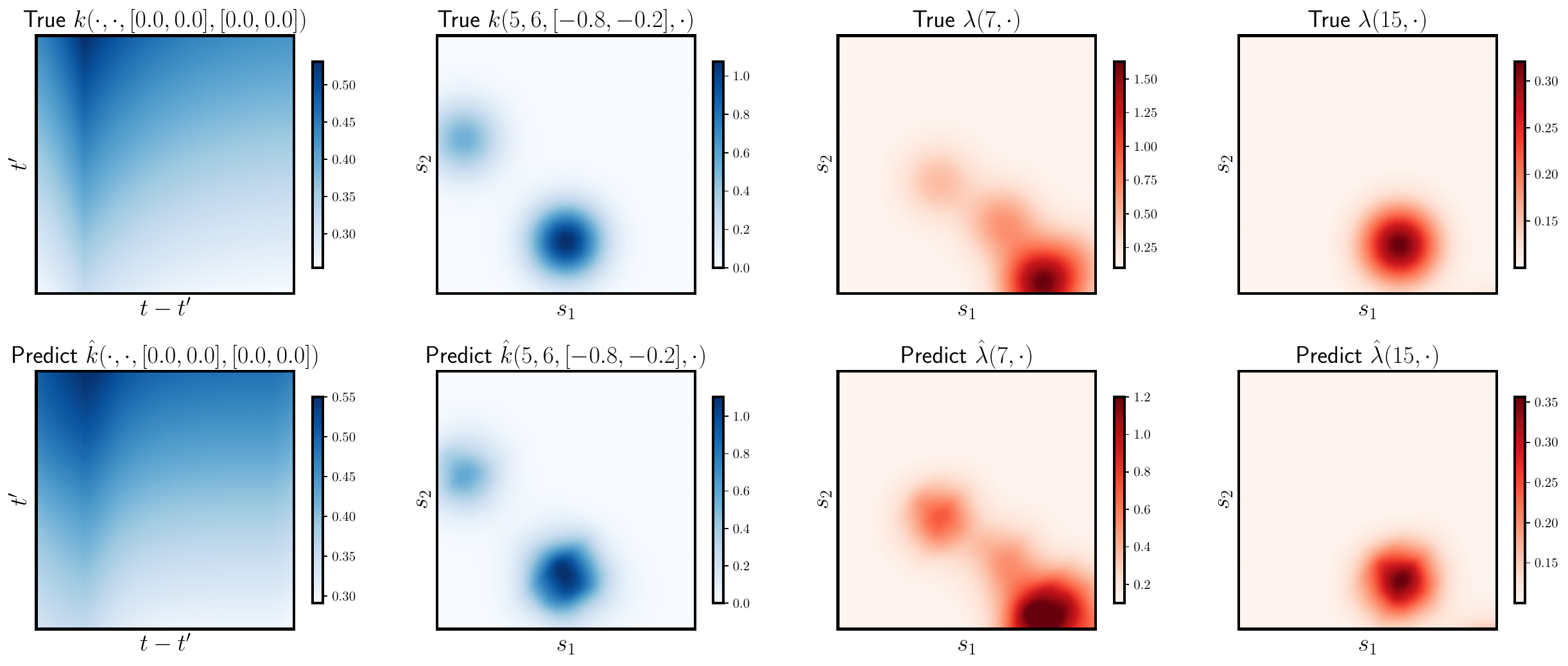}

\caption{
Kernel and intensity recovery results of a point process with a spatio-temporal non-stationary kernel. The top row shows the ground truth, and the bottom row shows the model learned by \texttt{DNSK}. The first two columns visualize the spatio-temporal propagation patterns of the true and learned kernels. The last two columns show snapshots of the true and predicted conditional intensity functions computed from a test sequence. The results suggest an accurate recovery of the ground truth by  \texttt{DNSK} and log-barrier to ensure nonnegativity of the intensity function.}
\label{fig:3d-data-experiment-2}
\end{figure}

This example demonstrates how a low-rank kernel can be estimated using MLE by a numerical optimization scheme of gradient descent and how well this recovers the true kernel. Specifically, we consider a non-stationary kernel of the form
\[
k\bigl(t^\prime, t, s^\prime, s\bigr) 
=\;
\sum_{r=1}^{2}\sum_{l=1}^{2}
\alpha_{rl}
\,u_r\!\bigl(s^\prime\bigr)
\,v_r\!\bigl(s - s^\prime\bigr)
\,\psi_l\!\bigl(t^\prime\bigr)
\,\varphi_l\!\bigl(t - t^\prime\bigr), \quad t'<t,
\]
where
\[
u_1\bigl(s^\prime\bigr) 
= 
1 - a_s\!\bigl(s_2^\prime + 1\bigr),
\quad
u_2\bigl(s^\prime\bigr) 
= 
1 - b_s\!\bigl(s_2^\prime + 1\bigr),
\quad
v_1\bigl(s - s^\prime\bigr) 
= 
\frac{1}{2\pi \sigma_1^2}
\,\exp\!\Bigl(-\frac{\lVert s - s^\prime\rVert^2}{2\sigma_1^2}\Bigr),
\]
\[
v_2\bigl(s - s^\prime\bigr) 
= 
\frac{1}{2\pi \sigma_2^2}
\,\exp\!\Bigl(-\frac{\lVert s - s^\prime - 0.8\rVert^2}{2\sigma_2^2}\Bigr),
\quad
\psi_1\!\bigl(t^\prime\bigr) 
= 
1 - a_t\,t^\prime,
\quad
\psi_2\!\bigl(t^\prime\bigr) 
= 
1 - b_t\,t^\prime,
\]
\[
\varphi_1\!\bigl(t - t^\prime\bigr) 
= 
\exp\!\bigl(-\beta\,(t - t^\prime)\bigr),
\quad
\varphi_2\!\bigl(t - t^\prime\bigr) 
= 
\bigl[t - t^\prime - 1\bigr]
\,\mathbb{I}\bigl(t - t^\prime < 3\bigr).
\]
The parameters are set to \(a_s = 0.3\), \(b_s = 0.4\), \(a_t = 0.02\), \(b_t = 0.02\), \(\sigma_1 = 0.2\), \(\sigma_2 = 0.3\), \(\beta = 2\), and
\[
(\alpha_{11},\, \alpha_{12},\, \alpha_{21},\, \alpha_{22})
=
(0.6,\, 0.15,\, 0.225,\, 0.525).
\]
The results are presented in {\bf Figure}~\ref{fig:3d-data-experiment-2}, where the kernel is modeled using \texttt{DNSK} \citep{dong2023spatio}, a representative deep-kernel method for spatio-temporal point processes. \textcolor{black}{While Figure 1 provides a schematic illustration, Figure 2 visualizes kernels actually learned by deep models, showing how they recover true underlying kernels.} Using maximum likelihood estimation (MLE) to estimate this low-rank kernel, we observe that the recovered kernel closely aligns with the ground truth. The figure shows both the estimated kernel and the predicted conditional intensity functions for a test sequence, demonstrating their close correspondence to the true values. These results confirm that the proposed approach effectively captures spatio-temporal dependencies and provides accurate intensity predictions when the underlying model is truly low-rank.

\subsubsection{Identifiability of kernel by MLE}  

A key theoretical question is whether the true kernel is {\it identifiable}, i.e., whether a unique maximum likelihood or least-squares solution exists in the large-sample regime. We investigate this question under the assumption that multiple event trajectories are observed.  
Consider \(M\) observed trajectories, each comprising event sequences \(\{x_{i,j}\}_{i=1}^{N_j}\) on the interval \([0, T]\). Here, each \(x_{i,j}\) represents a single event and may extend beyond the spatio-temporal setting. Let \(\lambda_j\) and \(\mathbb{N}_j\) respectively denote the conditional intensity and the counting measure for the \(j\)-th trajectory. The maximum likelihood estimation (MLE) for kernel recovery, while directly solving for the parameters of the kernel representation, can be interpreted as solving the following variational problem over kernels \(k\):
\begin{equation}
    \max_{k \in \mathcal{K}} 
    \ell[k] 
    \;\coloneqq\;
    \frac{1}{M} 
    \sum_{j=1}^M 
    \Bigl(
    \int_{\mathcal{X}} 
    \log \lambda_j[k](x) 
    \,d\mathbb{N}_j(x)
    \;-\;
    \int_{\mathcal{X}} 
    \lambda_j[k](x)\,dx
    \Bigr),
    \label{eq:log-likelihood}
\end{equation}
where \(\mathcal{K}\subset C^0(\mathcal{X} \times \mathcal{X})\) is the family of admissible kernel functions induced by the finite-rank decomposition in Equation \eqref{kernel} and the associated feature function family \(\mathcal{F}\).

Recent work  \cite{zhu2022ICLRneural} indicates that, under mild regularity assumptions (e.g., enforcing positivity constraints on the intensity), any small perturbation away from the true kernel reduces the objective, which guarantees identifiability. Let \(\overline{\mathcal{K}}\) be an extended kernel class containing \(\mathcal{K}\), allowing for possibly more general functions in \(C^0(\mathcal{X}\times \mathcal{X})\). Throughout the theory, the nonnegative spectrum can be subsumed into the feature functions for notational simplicity.

\begin{assumption}\label{assump:B1B2} 
\textbf{(A1)} \(\overline{\mathcal{K}}\subset C^0(\mathcal{X}\times \mathcal{X})\) is uniformly bounded, and the true kernel \(k^*\in \overline{\mathcal{K}}\).  
\textbf{(A2)} There exist positive constants \(c_1\) and \(c_2\) such that for any \(k \in \overline{\mathcal{K}}\), almost surely over each event trajectory, 
\[
c_1 \;\le\; \lambda[k](x) \;\le\; c_2, 
\quad 
\forall\, x \in \mathcal{X}.
\]
\end{assumption}
The following result is quoted from \cite{zhu2022ICLRneural}:
\begin{lemma}[Local perturbation of likelihood around the true kernel]
\label{lemma:perturb}
Under Assumption \ref{assump:B1B2}, for any \(\tilde{k}\in\overline{\mathcal{K}}\) and \(\delta k = \tilde{k} - k^*\),
\begin{equation}
\ell [k^*] 
\;-\;
\ell [\tilde{k}] 
\;\ge\;
 \frac{1}{M}
 \Bigl\{
 -\sum_{j=1}^M 
 \int_{\mathcal{X}} 
 \delta\lambda_j(x)
\Bigl(\frac{d\mathbb{N}_j(x)}{\lambda_j[k^*](x)}  - dx\Bigr)
\;+\;
\frac{1}{2 c_2^2 }
\sum_{j=1}^M 
\int_{\mathcal{X}} 
(\delta\lambda_j(x))^{2} 
\,d\mathbb{N}_j(x) 
\Bigr\},
\label{eq:perturb1}
\end{equation}
where
\begin{equation}\label{delta_lambda}
\delta \lambda_j(x) 
\;:=\; 
\int_{\mathcal{X}_{t(x)}} 
\delta k(x', x)
\,d \mathbb{N}_j(x').
\end{equation}
\end{lemma}

Lemma~\ref{lemma:perturb} shows that the change in likelihood caused by a perturbation around \(k^*\) decomposes into two terms in Equation \eqref{eq:perturb1}: (i) a martingale integral, whose expectation is zero because 
\(\mathbb{E}[d\mathbb{N}_j(x)/\lambda_j[k^*](x) - dx]=0\); and (ii) a quadratic term \(\int_{\mathcal{X}} (\delta \lambda_j(x))^2 d\mathbb{N}_j(x)\) that penalizes deviations in the intensity function. For each trajectory \(j\),
\[
\int_{\mathcal{X}}(\delta \lambda_j(x))^2\,d\mathbb{N}_j(x)
\;\approx\;
\int_{\mathcal{X}} (\delta \lambda_j(x))^2 \lambda_j^*(x)\,dx
\;\ge\;
c_1 \int_{\mathcal{X}} (\delta \lambda_j(x))^2\,dx,
\]
indicating that perturbations to \(\lambda_j\) reduce the log-likelihood.

\begin{theorem}[Kernel identifiability]
\label{main-thm}
Under Assumption \ref{assump:B1B2}, the true kernel \(k^*\) is locally identifiable in that \(k^*\) is a local maximizer of the likelihood \eqref{eq:log-likelihood} in expectation.
\end{theorem}

Theorem~\ref{main-thm} establishes kernel identifiability even within parametric families \(\mathcal{K}\) induced by feature functions \(\mathcal{F}\). In neural network–based kernels, parameter identifiability may be complicated by symmetries (e.g., permuting hidden neurons), yet the induced \emph{functional} identifiability remains valid under neural network approximation results. Hence, Theorem~\ref{main-thm} is significant for neural kernel learning: despite potential parameter-level degeneracies, the learned kernel function itself is identifiable.

\subsection{Least-Square (LS) model recovery}

An alternative approach for estimating \(\lambda\) is based on minimizing the discrepancy between the model intensity \(\lambda\) and the observed sample path of events, measured in terms of the Least Squares (LS) criterion. Although LS is less common in classical statistics, it can sometimes simplify the computational or theoretical analysis for neural-based kernels. Let \(d\mathbb{N}(t, s)\) denote the counting measure of events over time and space, and consider the set of observed events \(\{(t_i, s_i)\}_{i=1}^n\) within the time interval \([0, T]\) and a compact domain $s\in \mathcal S$. We write the model intensity function as \(\lambda(t, s)\), omitting the explicit dependence on the model parameters \(\theta\).

The population-level \(L^2\) loss can be expressed as
\begin{equation}
L_{\mathrm{LS}}
\,=\,
\mathbb{E}\!\int_{0}^{T}\!\!\int_{S}\!\Bigl[\lambda(t,s)^{2} 
- 2 \,\lambda(t,s)\,\lambda^*(t,s) \Bigr]
\,ds\,dt \label{pop_l2_loss}
\end{equation}
where \(\lambda^*(t,s)\,dt\,ds = \mathbb{E}\bigl[d\mathbb{N}(t,s)\,\big|\,\mathcal{H}_t\bigr]\), and \(\lambda(t,s) \in \mathcal{H}_t\); the population level \(L^2\) loss in Equation \eqref{pop_l2_loss} is equal to $\mathbb{E}\!\int_{0}^{T}\!\!\int_{S}\!\bigl[\lambda(t,s) - \lambda^*(t,s)\bigr]^{2}\,ds\,dt,$ up to an additive constant independent from model intensity $\lambda(t,s)$.

By definition and the property of conditional expectation,
\begin{align}
L_{\mathrm{LS}}
& = 
\mathbb{E}\! \int_{0}^{T}\!\!\int_{S}\!
\lambda(t,s)^{2} \,ds\,dt  
- \mathbb{E}\! \int_{0}^{T}\!\!\int_{S}\!
 2\,\lambda(t,s)\,\mathbb{E}\bigl(d\mathbb{N}(t,s)\,\big|\,\mathcal{H}_t\bigr)
\nonumber \\
& = 
\mathbb{E}\! 
\left[ \int_{0}^{T}\!\!\int_{S}\!
\lambda(t,s)^{2} \,ds\,dt  
- \! \int_{0}^{T}\!\!\int_{S}\!
 2\,\lambda(t,s)\, d\mathbb{N}(t,s)\,\right]. \nonumber
\end{align}
Thus, in practice, for a single trajectory with observations $(t_i, s_i)$, $i = 1, \ldots, n$,
we can compute the empirical loss as 
\begin{equation}\label{LS_loss}
 \ell_{\mathrm{LS}} = 
\int_{0}^{T}\!\!\int_{S}\!\lambda(t,s)^{2}\,ds\,dt
- 2\sum_{i=1}^{n}\lambda(t_i, s_i),
\end{equation}
and estimate the model parameter by minimizing the objective, i.e., solving \(\min_{\theta} \hat \ell_{\mathrm{LS}}(\theta)\). 
When there are multiple trajectories, one can form the LS loss in Equation \eqref{LS_loss} by averaging over multiple trajectories.

Comparing the LS objective in Equation \eqref{LS_loss} with the log-likelihood objective in Equation \eqref{eq:pp-log-likelihood}, we observe that the least squares (LS) objective does not involve a logarithmic term of the conditional intensity function \(\lambda(t, v)\),  which can improve numerical stability, particularly when the conditional intensity is close to zero. However, this also means that LS may insufficiently penalize small intensity values, potentially leading to suboptimal models in certain cases. Additionally, the evaluation of the integral over the space-time domain incurs a computational cost comparable to that of maximum likelihood estimation (MLE). Despite these limitations, the LS approach can demonstrate greater numerical stability and can yield strong empirical performance (see, e.g., \citet{dong2023deep}).

\section{Applications}\label{sec:app}

Modern self-exciting point processes combined with deep learning architectures have been widely applied to learn complex event dependencies across various domains. Notable applications of deep spatio-temporal point processes (STPPs) include:

\begin{itemize}
    \item \textit{Crime data modeling and contagious dynamics modeling of crime incidents}, as demonstrated in works such as \citet{mohler2011self, mohler2013modeling, zhu2022spatiotemporaltextual}, including studies on the impact of urban environments with street-network topology constraints and landmarks \citep{dong2024spatio}, as well as the effect of spatial covariates on crime intensity \citep{dong2024atlanta}.

    \item \textit{Flexible and scalable earthquake forecasting}, enabled by auto-regressive neural network outputs \citep{dascher2023using} or neural network-based kernels \citep{zhu2021imitation}.
    
    \item \textit{Health systems surveillance}, including clinical event prediction from timestamped interaction sequences \citep{enguehard2020neural}, deep kernel modeling of high-resolution infectious disease datasets with highly non-stationary spatio-temporal point patterns \citep{dong2023non}, and causal graph discovery for sepsis-associated derangements \citep{wei2023granger}.
    
\end{itemize}

This list is not exhaustive. Other notable applications include social network interactions \citep{zipkin2016point, li2017detecting}, traffic congestion prediction \citep{jin2023spatio, zhu2021spatio}, modeling civilian deaths in Iraq \citep{lewis2012self}, multiple object tracking \citep{wang2020spatio}, city taxi pick-up predictions \citep{okawa2019deep}, online advertisement \citep{xu2014path}, and football match event analysis \citep{yeung2023transformer}.

In the following sections, we present several real-data examples to illustrate the application of spatio-temporal point processes based on various deep kernels. \textcolor{black}{We briefly discuss the implications of each result, and some comparison of the deep kernel–based approach with alternative methods in Table \ref{fig:hd-data-experiment}, which highlights its promise. The examples are intended to illustrate how deep kernel–based approaches can be useful in these contexts. While the exposition reflects applications most familiar to the authors, related work by others is cited at the beginning of Section \ref{sec:app} to indicate that these applications are of broader interest. }

\subsection{Online prediction of earthquake events}

We demonstrate the model’s predictive capability by performing online prediction on earthquake data,  using  \texttt{DNSK} \citep{dong2023spatio}, which is based on the kernel expansion in Equation \eqref{eq:spatio-temporal-kernel-representation}, as a representative method using deep-kernel for the spatio-temporal process. The dataset, obtained from the open-source Southern California Earthquake Data Center (SCEDC) \citep{SCEDC}, contains time and location information for earthquakes in Southern California. We collect 19,414 earthquake records from 1999 to 2019 with magnitudes greater than 2.5 and partition the data into monthly sequences, each with an average length of 40.2 events. In this example, we focus on modeling the time and location of the earthquakes to illustrate the application of spatiotemporal modeling, although earthquake magnitude could also be incorporated for improved prediction.

For prediction, we compute the conditional probability that the next event occurs at \((t,s)\) given the history \(\mathcal{H}_t\) is
\[
    f(t, s) \;=\; \lambda(t, s)\,\exp\!\Bigl(-\!\!\int_{t_n}^{t}\!\!\int_{\mathcal{S}}\!\lambda(\tau, \nu)\,d\tau\,d\nu\Bigr).
\]
We then predict the time and location of the \((n+1)\)-th event via
\[
    \mathbb{E}\bigl[t_{n+1}\mid \mathcal{H}_t\bigr]
    \;=\;\int_{t_n}^{\infty}\!t \,\biggl(\int_{\mathcal{S}} f(t, s)\,ds\biggr)\!dt, 
    \quad
    \mathbb{E}\bigl[s_{n+1}\mid \mathcal{H}_t\bigr]
    \;=\;\int_{\mathcal{S}}\!s \,\biggl(\int_{t_n}^{\infty} f(t, s)\,dt\biggr)\!ds.
\]
We compute the mean absolute error (MAE) by comparing these predictions with the observed final event in each sequence, thereby quantifying the accuracy of the model’s time and location forecasts. The results are shown in {\bf Table} \ref{tab:real-data-results-transposed}:  \texttt{DNSK} provides more accurate predictions than other alternatives with higher event log-likelihood. \textcolor{black}{This example demonstrates the the deep kernel based approach has the potential to strike a balance between flexiblity of model, versus generalization power (as seen from the out-of-sample prediction capability), compared with the classical parametric kernel approach, and the influence kernel-less approach to directly model conditional intensity function modeling using deep neural networks. }

\begin{table}[!t]
  \caption{Real data results with Southern California Earthquake data. Testing log-likelihood (higher the better) and prediction mean absolute error of event time and location (lower the better) are reported. Reference for the methods compared can be found in {\bf Table} \ref{tab:literature-review}.}
  \vspace{.1in}
  \label{tab:real-data-results-transposed}
  \centering
  \resizebox{\textwidth}{!}{%
  \begin{tabular}{lcccccc}
    \toprule
    & \texttt{RMTPP} & \texttt{NH} & \texttt{THP} & \texttt{PHP+exp} & \texttt{NSMPP} & \texttt{DNSK} \\
    \midrule
    Testing $\ell$ & $-1.825_{(0.053)}$ & $-1.818_{(0.037)}$ & $-1.784_{(0.007)}$ & $-2.048_{(0.093)}$ & $-4.152_{(0.187)}$  & $\textbf{--1.751}_{(0.080)}$ \\
    Time MAE & $6.963$ & $6.880$ & $6.113$ & $8.132$ & $6.780$ & $\textbf{1.474}$ \\
    Location MAE & $0.602$ & $0.458$ & $0.633$ & $0.487$ & $0.455$  & $\textbf{0.431}$ \\
    \bottomrule
  \end{tabular}%
  }
\end{table}

\subsection{Atlanta police reports: Spatiotemporal data high-dimensional marks}
\label{sec:high-dim-mark}

 We demonstrate the use of deep kernel for spatio-temporal point processes with high-dimensional marks by adapting \texttt{DNSK} \citep{dong2023spatio} to consider high-dimensional marks.  We use proprietary crime data from the Atlanta Police Department (APD). This dataset contains 4,644 crime incidents from 2016 to 2017, each annotated with a timestamp, location, and a detailed text description. After applying TF-IDF, each event initially has a 5,012-dimensional text representation. Following \citet{zhu2022spatiotemporaltextual}, we map this high-dimensional, sparse representation into a 50-dimensional binary vector using a Restricted Boltzmann Machine (RBM) \citep{fischer2012introductionRBM}, which helps de-noise extraneous text features and enables summation-based computations in place of double integrals.

\begin{figure}[!tb]
\centering

\begin{subfigure}[h]{\linewidth}
\includegraphics[width=\linewidth]{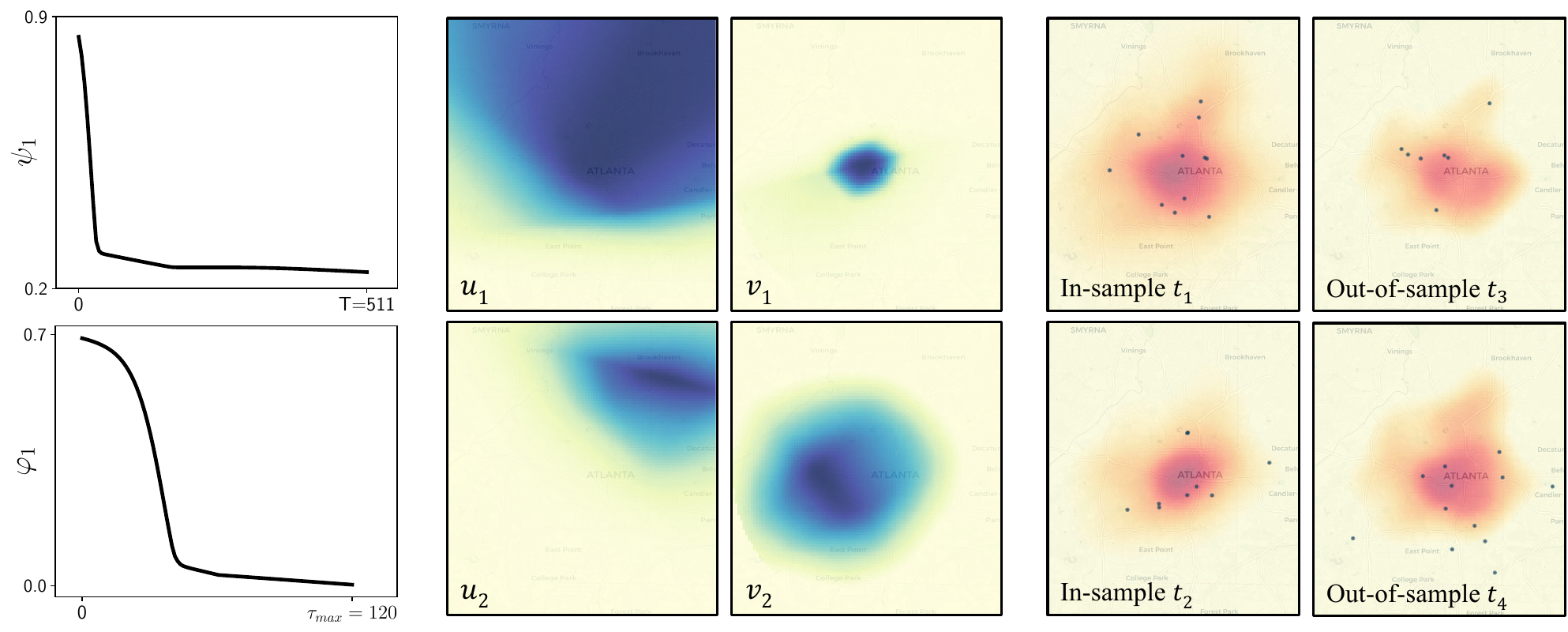}
\end{subfigure}

\caption{Model fitting and prediction for police data with text (high-dimensional marks). The first column shows the learned temporal functions. The four panels in the middle display the learned spatial functions. Note that we do not specify the shape of the temporal influence kernel; instead, the basis functions parameterized by a neural network are able to capture the temporal decaying pattern, including its shape. Darker colors indicate higher function values. The last four panels show the predicted conditional intensity over space at two in-sample time points and two out-of-sample time points. The dots represent the event occurrences on each corresponding day.
}
\label{fig:hd-data-experiment}
\end{figure}

{\bf  Figure}~\ref{fig:hd-data-experiment} illustrates the learned influence-kernel basis functions, revealing a decaying temporal effect and distinct spatial influence patterns, particularly in the northeast region. The in-sample and out-of-sample intensity predictions also confirm the model's ability to capture variations in event occurrence by adapting its conditional intensity accordingly. \textcolor{black}{This example demonstrates the capability of deep kernels in handling complex high-dimensional marks (in this case, free text in police reports). This simple demonstration may not fully investigate the advantages and limitations of deep kernel–based approaches compared with their alternatives. Still, its out-of-sample prediction performance shows considerable promise and points to an avenue for future research.}

\subsection{Example: Graph-based deep kernel learning dynamic graph for sepsis data}
\label{sec:GNN_dp}

In this example, we demonstrate the use of a deep kernel method based on GNN for graph filtering, called \texttt{GraDK}, proposed in \citep{dong2023deep} and based on the kernel expansion in Equation \ref{eq:temporal-graph-decomposed-kernel}. The data we consider is ICU time-series data for sepsis prediction to demonstrate the model’s capability in learning dynamic graph influence, where each node in the graph corresponds to a medical variable. The Sepsis dataset, released by the PhysioNet Challenges \citep{goldberger2000physiobank, reyna2019early}, is a physiological dataset that records patient covariates—including demographics, vital signs, and laboratory values—for ICU patients from three separate hospital systems. Based on these covariates, common and clinically relevant Sepsis-Associated Derangements (SADs) can be identified through expert clinical judgment. A SAD is considered present when a patient’s covariates fall outside normal limits \citep{wei2023granger}. In our analysis, we extract the onset times of 12 SADs and sepsis (a total of 13 medical indices) as discrete events, resulting in a total of 80,463 events. Each sequence corresponds to a single patient’s events over a 24-hour period. Event times are measured in hours, with the average sequence length being 15.6 events.

\begin{figure}[!t]
\vspace{-0.05in}
    \centering
    \begin{subfigure}[h]{.95\linewidth}
    \includegraphics[width=1\linewidth]{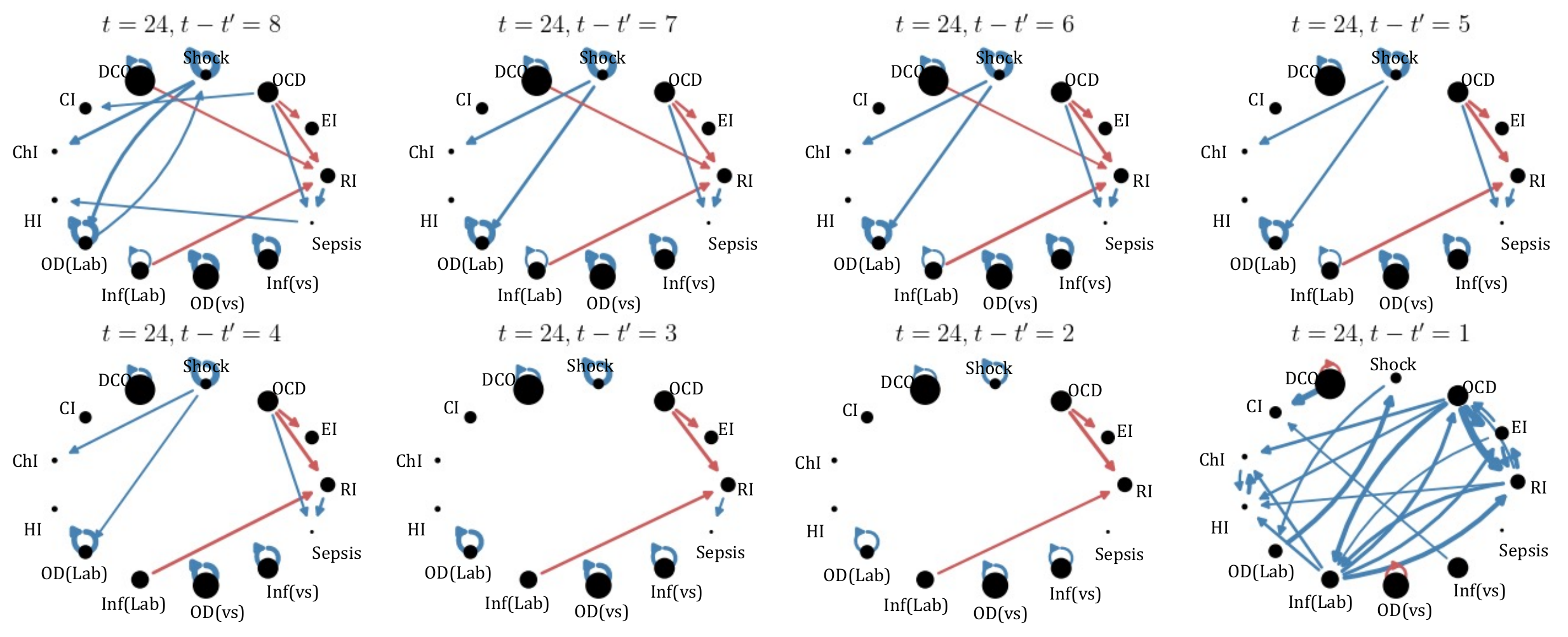}
    \end{subfigure}
    \vspace{-0.05in}
\caption{Dynamic graph influence learned by \texttt{GraDK} on the Sepsis data. The learned kernel is evaluated at eight time lags from $1$ to $8$. The node radii are proportional to the background intensity on each node, and the edge widths are proportional to the influence magnitude. Blue and red edges represent the excitation and inhibition effects, respectively. The current time $t$ is fixed to be $24$ (the last hour) for all panels. 
}
\label{fig:sepsis-dynamic-kernel}
\end{figure}

For such a problem, we estimate the kernel to $\hat k$, using data; the meaning of the learning kernel is that it captures when the time gets close to the sepsis onset, how the different medical variables influence each other and cause ``sepsis'' eventually, changes over time. 
{\bf Figure}~\ref{fig:sepsis-dynamic-kernel} visualizes eight snapshots of the influence kernel $\hat k(t, t', v, v')$ learned by \texttt{GraDK} among different medical variables at different lags $t-t'$. We visualize $k(t, t', \cdot, \cdot)$ by treating it as the incident matrix of a directed graph. 
The snapshots are arranged in an order with the decrease of the time lag $t-t'$, showcasing the temporal evolution of the graph influence. The results reveal a decaying temporal pattern in the interactions across indices. In particular, the onset of Sepsis is excited by the earlier occurrences of certain medical indices (e.g., RI, OCD, DCO, OD(vs), Inf(vs)), with these excitation effects being most prominent at early time points and gradually fading over time. Overall, these medical indices—excluding Sepsis itself—tend to exhibit stronger interactions when Sepsis does not eventually occur. \textcolor{black}{This example demonstrates the capability of the deep graph kernel–based approach in capturing dynamic graphs, which highlights its potential for interpreting the learned kernel. Its connection to dynamic causal graph learning presents a promising direction for future research..}

\section{Conclusion and Outlook}

Deep non-stationary kernels for spatio-temporal point processes (STPPs) have opened new avenues for modeling complex event data by relaxing restrictive parametric assumptions and leveraging the strong representational power of neural networks, while preserving the explainability of results through influence kernels. Their ability to capture intricate temporal and spatial dependencies—in domains such as crime analysis, earthquake aftershocks, and social network cascades—has led to improvements in both predictive accuracy and interpretability. Recent developments also provide theoretical guarantees for identifiability and offer efficient computational strategies.

Despite these advances, several challenges remain. First, scaling to high-dimensional or massive event streams necessitates more efficient training algorithms and strategies to mitigate overfitting. Second, interpretability can be enhanced through regularization or domain-informed constraints, striking a balance between flexible model representations and clarity in the learned mechanisms. Third, identifiability issues—especially distinguishing between time- or location-varying baseline intensities and event-triggered effects—require further theoretical scrutiny. Incorporating prior knowledge into kernel design may help address these concerns and lead to both theoretical and practical improvements.

Emerging directions promise to further extend the capabilities of neural STPPs. Graph-based kernels enriched by graph neural networks enable event modeling on complex relational structures. Causal discovery frameworks can help uncover underlying mechanisms and interactions from learned kernels, while advanced generative methods—particularly diffusion- or score-based approaches—offer alternatives to intensity-function-based models that may improve scalability for large-scale and multi-dimensional data.

It is also worth mentioning that the deep kernel approach discussed here—as well as most existing work—focuses on modeling the influence of past events as being additive. Extensions to multiplicative influence effects \citep{duval2022interacting} or other types of interactions \citep{perry2013point} represent promising directions for future research.

Another important topic is uncertainty quantification, which is crucial for both kernel estimation and subsequent predictions, given the stochastic nature of event occurrence. Techniques for modeling and propagating uncertainty within deep STPPs \citep[e.g.,][]{wang2020uncertainty} could enhance robustness and confidence in model outputs. Finally, systematic frameworks for causal inference and interpretability—rooted in both classical statistics and modern machine learning—are poised to elevate the impact of neural STPPs across domains such as seismology, public health, and security. Robust collaboration among statisticians, computer scientists, and domain specialists will be essential for realizing the full potential of STPPs to drive scientific insight and informed decision-making.

\section*{DISCLOSURE STATEMENT}

The authors are not aware of any affiliations, memberships, funding, or financial holdings that
might be perceived as affecting the objectivity of this review.

\section*{ACKNOWLEDGMENTS}

Y. X. is partially supported by NSF DMS-2134037, CNS-2220387, and the Coca-Cola Foundation. X. C. is also partially supported by NSF DMS-2237842
and Simons Foundation MPS-MODL-00814643.

\bibliography{arxiv_refs,jcgs_refs,refs}
\bibliographystyle{apalike}

\end{document}